\documentclass{article} 
\usepackage[usenames,dvipsnames]{color}
\usepackage{nips14submit_e,times}
\usepackage[T1]{fontenc}
\usepackage{lmodern}
\usepackage{natbib}
\usepackage{latexsym}
\usepackage{amsmath}
\usepackage{multirow}
\usepackage{graphicx}
\usepackage{amsfonts,eucal,amsbsy}
\usepackage{amsmath}
\usepackage{amssymb}
\usepackage{sidecap}
\usepackage{microtype}
\usepackage{mdwlist}
\usepackage{algorithm}
\usepackage{algorithmic}
\usepackage{subfigure}
\usepackage{hyperref}
\usepackage{url}

\usepackage[font=small]{caption}
\usepackage{sidecap}

\usepackage{enumitem}

\setitemize{noitemsep,topsep=6pt,parsep=0pt,partopsep=0pt,itemindent=0pt,leftmargin=12pt,labelwidth=12pt}
\setenumerate{noitemsep,topsep=6pt,parsep=0pt,partopsep=0pt}

\makeatletter
\renewcommand{\paragraph}{%
  \@startsection{paragraph}{4}%
  {\z@}{0ex \@plus 0ex \@minus 0ex}{-1em}%
  {\normalfont\normalsize\bfseries}%
}
\makeatother

\DeclareMathOperator*{\argmin}{arg\,min}

\newcommand{\vect}[1]{\mathbf{#1}}

\newcommand{\pen}{\Omega}

\newcommand{\thisapproach}{\textsc{forest}}
\newcommand{\littleparen}[1]{{\tiny(#1)}}

\title{Learning Word Representations \\ with Hierarchical Sparse Coding}

\author{Dani Yogatama \quad Manaal Faruqui \quad Chris Dyer \quad Noah A. Smith \\
Language Technologies Institute \\
Carnegie Mellon University \\
Pittsburgh, PA 15213, USA \\
\texttt{\{dyogatama,mfaruqui,cdyer,nasmith\}@cs.cmu.edu}
}

\nipsfinalcopy 

\begin{document}
\maketitle

\begin{abstract}
We propose a new method for learning word representations using
hierarchical regularization in sparse coding inspired by the linguistic study
of word meanings.
We show an efficient learning algorithm based on stochastic proximal methods
that is significantly faster than previous approaches, 
making it possible to perform hierarchical sparse coding on a corpus
of billions of word tokens. 
Experiments on various benchmark tasks---word similarity ranking,
analogies, sentence completion, and sentiment analysis---demonstrate
that the method outperforms or is competitive with state-of-the-art methods.
Our word representations are available at \url{http://www.ark.cs.cmu.edu/dyogatam/wordvecs/}.
\end{abstract}

\section{Introduction}
When applying machine learning to text, the classic
categorical representation of words as indices of a vocabulary fails
to capture syntactic and semantic similarities that are easily
discoverable in data (e.g., \emph{pretty}, \emph{beautiful},
and \emph{lovely} have similar meanings, opposite to
\emph{unattractive}, \emph{ugly}, and \emph{repulsive}).  
In contrast, recent
approaches to word representation learning apply neural networks to
obtain dense, low-dimensional, continuous embeddings of words
\citep{nlm,mnihicml,collobert,huang,mikolovrnn,mikolovsg,hpca}. 

In this work, we propose an alternative approach 
 based on decomposition of a high-dimensional matrix
capturing surface statistics of association between a word and
its ``contexts'' with sparse coding.  As in past work, contexts are words that occur nearby in running text
\citep{turney}. 
Learning is performed by minimizing a reconstruction loss function to 
find the best factorization of the input matrix.

The key novelty in our method is to govern the relationships among
dimensions of the learned word vectors, introducing a hierarchical
organization imposed through a structured penalty known as the group lasso \citep{yuan}. 
The idea of regulating the order in which variables enter a model was first proposed by \citet{binyu},
and it has since been shown useful for other applications \citep{hierarchicalsparsecoding}.
Our approach is motivated by coarse-to-fine organization of words' meanings 
often found in the field of lexical semantics 
(see \S{\ref{sec:structreg}} for a detailed description),
which mirrors evidence for distributed nature of hierarchical concepts
in the brain
\citep{raposo}.  Related ideas have also been explored in syntax \citep{petrov-klein:2008:EMNLP}.
It also has a foundation in cognitive science, where
hierarchical structures have been proposed as representations of
semantic cognition \citep{cogsci}.
We show a stochastic proximal algorithm for hierarchical sparse coding
that is suitable for problems where the input matrix is very large and sparse.
Our algorithm enables application of hierarchical sparse coding to
learn word representations from a corpus of billions of
word tokens and 400,000 word types. 

On standard evaluation tasks---word similarity ranking, analogies,
sentence completion, and sentiment analysis---we find that our method
outperforms or is competitive with the best published representations. 
 
\section{Model}
\subsection{Background and Notation}
The observable representation of word $v$ is taken to be a vector
$\vect{x}_v \in \mathbb{R}^C$ of
cooccurrence statistics with $C$ different contexts.  Most commonly,
each context is a possible neighboring word within a fixed
window.\footnote{Others include: global context \citep{huang},
  multilingual context \citep{manaal}, geographic context
  \citep{bamman-14}, brain activation data \citep{alona}, and
  second-order context \citep{schutze}.}
Following many others, we
let $x_{v,c}$ be the pointwise mutual information (PMI) between
the occurrence of context word $c$ within a five-word window of an occurrence
of word $v$ \citep{turney, nnse,manaal}.

In sparse coding, the goal is to represent each input vector
$\vect{x} \in \mathbb{R}^C$ as a sparse linear combination of
basis vectors. 
Given a stacked input matrix $\vect{X} \in \mathbb{R}^{C \times V}$, where $V$ is the number of
words, we seek to minimize:
\begin{align}
\label{eq:loss}
\argmin_{\vect{D} \in \mathcal{D},\vect{A}} \Vert \vect{X} - \vect{D}\vect{A} \Vert_2^2 + \lambda \pen(\vect{A}),
\end{align}
where $\vect{D} \in \mathbb{R}^{C \times M} $ is the dictionary of basis vectors,
$\mathcal{D}$ is the set of matrices whose columns have small (e.g.,
less than or equal to one) $\ell_2$ norm,
$\vect{A} \in \mathbb{R}^{M \times V}$ is the code matrix,
$\lambda$ is a regularization hyperparameter, and $\pen$ is the
regularizer.
Here, we use the squared loss for the reconstruction error, but other loss functions could also be used \citep{honglak}.
Note that it is not necessary, although typical,
for $M$ to be less than $C$ (when $M > C$, it is often called an overcomplete representation).
The most common regularizer is the $\ell_1$ penalty, which results in
sparse codes.  While structured regularizers are associated with
sparsity as well (e.g., the group lasso encourages group sparsity),
our motivation is to use $\Omega$ to encourage a coarse-to-fine organization of latent
dimensions of the learned representations of words.

\subsection{Structured Regularization for Word Representations}
\label{sec:structreg}

\begin{figure*}
\centering
\includegraphics[scale=0.6]{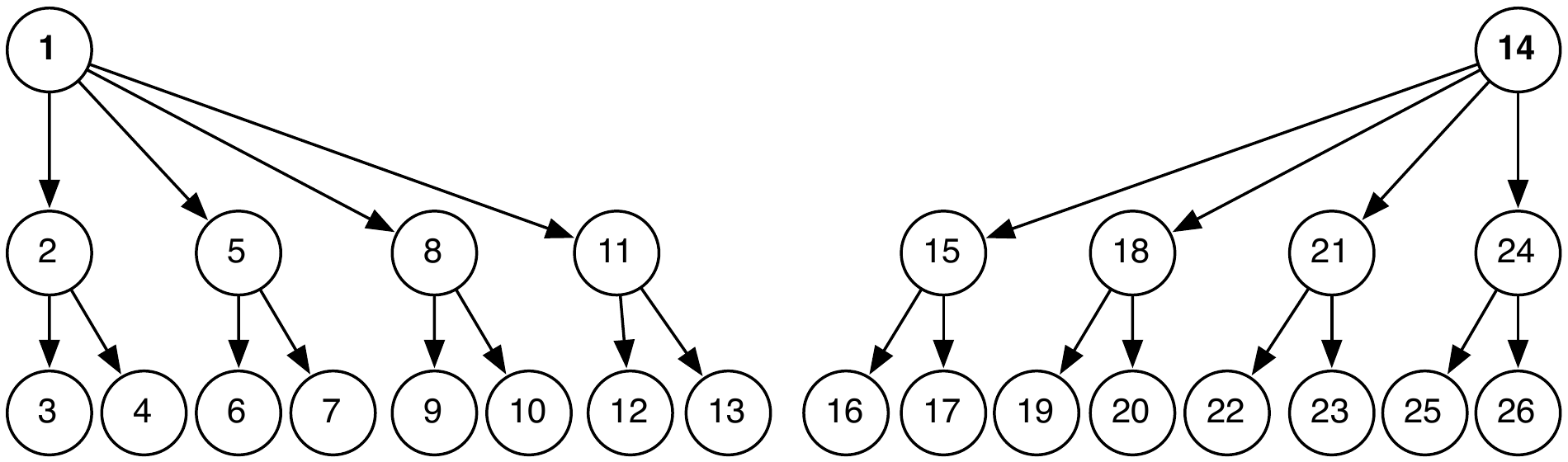}
\caption{An example of a regularization forest that governs the order in which variables enter the model.
In this example, $1$ needs to be selected (nonzero) for $2,3,\ldots,13$ to be selected.
However, $1,2,\ldots,13$ have nothing to do with the variables in the second tree: $14,15,\ldots,26$.
See text for details.
\label{fig:concept-tree}}
\end{figure*}

For $\Omega(\vect{A})$,
we design a forest-structured regularizer that encourages the model to use some
dimensions in the code space before using other dimensions.
Consider the trees in Figure~\ref{fig:concept-tree}.
In this example, there are 13 variables in each tree, and 26 variables
in total (i.e., $M=26$), each corresponding to a latent dimension for
\emph{one particular word}. 
These trees describe the order in which variables  ``enter the
model'' (i.e., take nonzero values).  
In general, a node may take a nonzero value only if its ancestors also
do.
For example, nodes 3 and 4 may only be nonzero  if nodes 1 and 2 are
also nonzero.
Our regularizer for column $v$ of $\vect{A}$, 
denoted by $\vect{a}_v$ (in this example, $\vect{a}_v \in \mathbb{R}^{26}$), for the trees in Figure~\ref{fig:concept-tree} is:
\begin{align*}
\pen( \vect{a}_v ) =&  \sum_{i=1}^{26} \Vert \langle a_{v,i}, \ \ 
\vect{a}_{v,\mathrm{Descendants}(i)} \rangle \Vert_2
\end{align*}
where $\mathrm{Descendants}(i)$ returns the (possibly empty) set of
descendants of node $i$.
\citet{hierarchicalsparsecoding} proposed a related penalty with only one tree
for learning image and document representations. 

Let us analyze why organizing the code space this way is helpful in learning better word representations.
Recall that the goal is to have a good dictionary $\vect{D}$ and code matrix $\vect{A}$.
We apply the structured penalty to each column of $\vect{A}$.
When we use the same structured penalty in these columns,
we encode an additional shared constraint that
 the dimensions of $\vect{a}_v$ that correspond to top level nodes should focus on ``general''
contexts that are present in most words.
In our case, this corresponds to contexts with extreme PMI values for most words,
since they are the ones that incur the largest losses. 
As we go down the trees, more word-specific contexts can then be captured. 
As a result, we have better organization \emph{across} words when learning their representations,
which also translates to a more structured dictionary $\vect{D}$. 
Contrast this with the case when we use unstructured regularizers 
that penalize each dimension of $\vect{A}$ independently (e.g., lasso).
In this case, each dimension of $\vect{a}_v$ has more 
flexibility to pay attention to any contexts 
(the only constraint that we encode is that the cardinality of the model should be small).
We hypothesize that this is less appropriate for learning word representations, since
the model has excessive freedom when learning $\vect{A}$ on noisy PMI values,
which translates to poor $\vect{D}$. 

The intuitive motivation for our regularizer comes from the field of
lexical semantics, which often seeks to capture the relationships between
words' meanings in  hierarchically-organized lexicons.  
The best-known example is WordNet \citep{wordnet}.
Words with the
same (or close) meanings are grouped together (e.g., \emph{professor}
and \emph{prof} are synonyms), and fine-grained
meaning groups (``synsets'') are nested under coarse-grained ones
(e.g., \emph{professor} is a hyponym of \emph{academic}). 
Our hierarchical sparse coding approach is still several steps away from
inducing such a lexicon, but it seeks to employ the dimensions of a
distributed word representation scheme in a similar coarse-to-fine
way.
In cognitive science, such hierarchical organization of
semantic representations was first proposed by \citet{cogsci}.

\subsection{Learning}
\label{sec:learning}
Learning is accomplished by minimizing the function in Eq.~\ref{eq:loss},
with the group lasso regularization function described in \S{\ref{sec:structreg}}.
The function is not convex with respect to $\vect{D}$ and $\vect{A}$,
but it is convex with respect to each when the other is fixed.
Alternating minimization routines have been shown to work reasonably
well in practice for such problems \citep{honglaktrain}, but they are
too expensive here due to:
\begin{itemize} 
\item The size of $\vect{X} \in \mathbb{R}^{C\times V}$ ($C$ and $V$ are
each on the order of $10^5$).
\item The many overlapping groups in
the structured regularizer $\Omega(\vect{A})$.
\end{itemize}

One possible solution is based on 
the online dictionary learning method of \citet{mairal}.  For $T$
iterations, we:
\begin{itemize}
\item Sample a mini-batch of words and (in parallel) solve for each one's $\vect{a}$ using the
alternating directions method of multipliers, shown to work well for
overlapping group lasso problems \citep{qin,yogatama2014a}.\footnote{Since our groups form tree structures, other methods such as FISTA \citep{hierarchicalsparsecoding} could also be used.}
\item Update $\vect{D}$ using the block coordinate descent algorithm
of \citet{mairal}.  
\end{itemize}
Finally,  we
parallelize solving for all columns of $\vect{A}$, which are separable
once $\vect{D}$ is fixed. 
In our experiments, we use this algorithm for a medium-sized corpus.

The main difficulty of learning word representations with hierarchical sparse coding is
that the size of the input matrix can be very large. 
When we use neighboring words as the contexts, the numbers of rows and columns are the size of the vocabulary.
For a medium-sized corpus with hundreds of millions of word tokens,
we typically have one or two hundred thousand unique words,
so the above algorithm is still applicable. 
For a large corpus with billions of word tokens, this number can easily double or triple, making
learning very expensive.
We propose an alternative learning algorithm for such cases.

We rewrite Eq.~\ref{eq:loss} as:
\begin{equation*}
\argmin_{\vect{D},\vect{A}} \sum_{c,v} (x_{c,v} - \vect{d}_c\cdot\vect{a}_v)^2 + \lambda \pen(\vect{A}) + \tau \sum_m \Vert \vect{d}_m \Vert_2^2
\end{equation*}
where (abusing notation) $\vect{d}_c$ denotes
the $c$-th row vector of $\vect{D}$ and $\vect{d}_m$ denotes
the $m$-th column vector of $\vect{D}$ (recall that $\vect{D} \in \mathbb{R}^{C\times M}$).
Instead of considering all elements of the input matrix, our algorithm approximates
the solution by using only non-zero entries 
in the input matrix $\vect{X}$. 
At each iteration, we sample a non-zero entry $x_{c,v}$
and perform gradient
updates to the corresponding row $\vect{d}_c$ and column $\vect{a}_v$.

We directly penalize columns of $\vect{D}$ by their squared $\ell_2$ norm
as an alternative to constraining columns of $\vect{D}$ to have unit $\ell_2$
norm.
The advantage of this transformation is that we have eliminated 
a projection step for columns of $\vect{D}$.
Instead, we can include the gradient of the penalty term in the stochastic gradient update.
We apply the proximal operator associated with $\pen(\vect{a}_v)$
as a composition of elementary proximal operators with no group overlaps, similar to \citet{hierarchicalsparsecoding}.
This can be done by recursively visiting each node of a tree and applying the proximal operator
for the group lasso penalty associated with that node (i.e., the group lasso penalty where the node
is the topmost node and the group consists of the node and all of its descendants).
The proximal operator associated with node $m$, denoted by $\text{prox}_{\Omega_m,\lambda}$,
is simply the block-thresholding operator for node $m$ and all its descendants.

Since each non-zero entry $x_{c,v}$ only depends on $\vect{d}_c$ and $\vect{a}_v$,
we can sample multiple non-zero entries and perform the updates in parallel as
long as they do not share $c$ and $v$.
In our case, where $C$ and $V$ are on the order of hundreds of thousands and we only have tens or hundreds of processors, 
finding non-zero elements that do not violate this constraint is easy.
There are typically a huge number of non-zero entries (on the order of billions).
Using a sampling procedure that favors entries with higher (absolute) PMI values
can lead to reasonably good word representations faster.
We sample a non-zero entry with probability proportional
to its absolute value.
This also justifies using only the non-zero entries,
since the probability of sampling zero entries is always zero.\footnote{In practice, we can use a faster approximation of this sampling procedure by uniformly sampling a non-zero entry and
multiplying its gradient 
by a scaling constant proportional to
its absolute PMI value.}
We summarize our learning algorithm in Algorithm~\ref{alg:proxgrad}.

\begin{algorithm*}[t]
   \caption{Fast algorithm for learning word representations with the forest regularizer.}
   \label{alg:proxgrad}
\begin{algorithmic}
   \REQUIRE matrix $\vect{X}$, regularization constant $\lambda$ and $\tau$, learning rate sequences $\eta_0,\ldots,\eta_T$, number of iterations $T$
   \STATE Initialize $\vect{D}_0$ and $\vect{A}_0$ randomly
   \FOR{$t=1,\ldots,T$ \{can be parallelized, see text for details\}}
   \STATE Sample $x_{c,v}$ with probability proportional to its (absolute) value
   \STATE $\vect{d}_c = \vect{d}_c + 2\eta_{t}  (\vect{a}_v (x_{c,v} -
   \vect{d}_c\cdot\vect{a}_v) - \tau \vect{d}_c)$
   \STATE $\vect{a}_v = \vect{a}_v +  2\eta_{t}  (\vect{d}_c (x_{c,v}
   - \vect{d}_c\cdot\vect{a}_v))$
   \FOR{$m=1,\ldots,M$}
   \STATE $\text{prox}_{\pen_m,\lambda}(\vect{a}_{v})$, where $\pen_m = \Vert\langle a_{v,m}, \ \ \vect{a}_{v,\mathrm{Descendants}(m)}\rangle \Vert_2$
   \ENDFOR
   \ENDFOR
\end{algorithmic}
\end{algorithm*}

\section{Experiments}

We present a controlled comparison of the forest regularizer against
several strong baseline word representations learned on a fixed
dataset, across several tasks.  In \S\ref{se:found} we compare to
publicly available word
vectors trained on different data.

\subsection{Setup and  Baselines}
We use the WMT-2011 English news corpus as our training data.\footnote{\url{http://www.statmt.org/wmt11/}}
The corpus contains about 15 million sentences and 370 million words. 
The size of our vocabulary is 180,834.\footnote{
We replace words with frequency less than 10 with \#rare\#
and numbers with \#number\#.
}

In our experiments, we use forests similar to those in Figure~\ref{fig:concept-tree} to organize the latent word space.
Note that the example has 26 nodes (2 trees). 
We choose to evaluate performance with $M=52$ (4 trees) and $M=520$
(40 trees).\footnote{In preliminary experiments we explored binary tree
structures and found they did not work as well; we leave a more
extensive exploration of tree structures to future work.}
We denote the sparse coding method with regular $\ell_1$ penalty by SC,
and our method with structured regularization
(\S{\ref{sec:structreg}}) by \thisapproach{}. 
We set $\lambda = 0.1$. 
In this first set of experiments with a medium-sized corpus, we use the online learning algorithm of \citet{mairal}.

We compare with the following baseline methods: 
\begin{itemize}
\item \citet{turney}: principal component analysis (PCA) by truncated singular value decomposition on $\vect{X}^{\top}$.
Note that this is also the same as minimizing the squared
reconstruction loss in Eq.~\ref{eq:loss} without any penalty on
$\vect{A}$.
\item \citet{mikolovrnn}: a recursive neural network (RNN) language model. 
We obtain an implementation from \url{http://rnnlm.org/}.
\item \citet{mnihicml}: a log bilinear model that predicts a word given its context, trained using noise-contrastive estimation (NCE, \citealp{nce}). We use our own implementation for this model.
\item \citet{mikolovsg}: a log bilinear model that predicts a word given its context (continuous bag of words, CBOW), trained using negative sampling \citep{mikolovneg}. We obtain an implementation from \url{https://code.google.com/p/word2vec/}.
\item \citet{mikolovsg}: a log bilinear model that predicts context words given a target word (skip gram, SG), trained using negative sampling \citep{mikolovneg}. We obtain an implementation from \url{https://code.google.com/p/word2vec/}.
\end{itemize}

Our focus here is on comparisons of model architectures.
For a fair comparison, we train all competing methods on the same
corpus using a context window of five words (left and right).
For the baseline methods, we use default settings in the provided implementations 
(or papers, when implementations are not available and we reimplement the methods).
We also trained the last two baseline methods with hierarchical softmax using a binary Huffman tree instead of negative sampling; 
consistent with \citet{mikolovneg}, we found that negative sampling
performs better and relegate hierarchical softmax results to
supplementary materials.

\subsection{Evaluation}
We evaluate on the following benchmark tasks.

\paragraph{Word similarity} The first task evaluates how well the representations capture word similarity.
For example \emph{beautiful} and \emph{lovely} should be closer in distance than
\emph{beautiful} and \emph{unattractive}.  We evaluate on a suite of
word similarity datasets, subsets of which have been considered in
past work:  WordSim 353 \citep{finkelstein}, rare words
\citep{luong}, and many others; see supplementary materials for details.
Following standard practice, for each competing model,
we compute cosine distances between word pairs in word similarity
datasets, then rank and report
Spearman's 
rank correlation coefficient \citep{spearman} between the model's rankings and human rankings.

\paragraph{Syntactic and semantic analogies} The second evaluation dataset is two analogy tasks proposed by \citet{mikolovsg}.
These questions evaluate syntactic and semantic relations between words.
There are 10,675 syntactic questions (e.g., \emph{walking : walked ::
  swimming : swam}) and
8,869 semantic questions (e.g., \emph{Athens : Greece :: Oslo ::
  Norway}).  
In each question, one word is missing, and the task is to correctly
predict the missing word.
We use the vector offset method 
\citep{mikolovsg} that computes the vector 
$\vect{b} = \vect{a}_{\text{Athens}} - \vect{a}_{\text{Greece}} 
+ \vect{a}_{\text{Oslo}}$.
We only consider a question to be answered correctly if
the returned vector ($\vect{b}$) has the highest cosine similarity to
the correct answer (in this example, $\vect{a}_{\text{Norway}}$). 

\paragraph{Sentence completion} The third evaluation task is the Microsoft Research sentence completion challenge \citep{msrsentence}.
In this task, the goal it to choose from a set of five candidate words
which one best completes a sentence.  For example:
\emph{Was she his \{client, musings, discomfiture, choice, opportunity\}, his friend, or his mistress?}
 (\emph{client} is the correct answer).
We choose the candidate with the highest average similarity to every other word in the sentence.\footnote{We
note that unlike matrix decomposition based approaches, some of the neural network based models can directly compute the scores of
context words given a possible answer \citep{mikolovsg}.
We choose to use average similarities for a fair comparison of the
representations.}

\paragraph{Sentiment analysis} The last evaluation task is sentence-level sentiment analysis.
We use the movie reviews dataset from \citet{socher}.
The dataset consists of 6,920 sentences for training, 872 sentences for development,
and 1,821 sentences for testing.
We train $\ell_2$-regularized logistic regression to predict binary sentiment, tuning the
regularization strength on development data.
We represent each example (sentence) as an $M$-dimensional vector
constructed by taking the average of word representations of words appearing in that sentence.

The analogy, sentence completion, and sentiment analysis tasks are
evaluated on prediction accuracy.

\subsection{Results}
Table~\ref{tbl:allresults} shows results on all evaluation tasks for $M=52$ and $M=520$.
Runtime will be discussed in \S{\ref{sec:discussion}}.
In the similarity ranking and sentiment analysis tasks, our method performed the best in both low and high dimensional embeddings.
In the sentence completion challenge, our method performed best in the high-dimensional case
and second-best in the low-dimensional case.
Importantly, \thisapproach{} outperforms PCA and
unstructured sparse coding (SC) on every task.
We take this collection of results as support for the idea that
coarse-to-fine organization of latent dimensions of word representations 
captures the relationships between words' meanings better compare to
unstructured organization.

\begin{table*}[t]
\caption{
Summary of results.
We report Spearman's correlation coefficient for the word similarity
task and accuracies (\%) for other tasks.
Higher values are better (higher correlation coefficient or higher accuracy).
The last two methods (columns) are new to this paper, and our proposed
method is in  the last column.
\label{tbl:allresults}
}
\centering 
\begin{tabular}{|c|l||r|r|r|r|r||r|r|}
\hline
{$M$} & {Task} & {PCA} & {RNN}& {NCE} &{CBOW} & {SG} 
& {SC}& \thisapproach{}\\
\hline
\hline
\multirow{5}{*}{52} & Word similarity &  0.39 & 0.26& 0.48 & 0.43 & 0.49
& 0.49 & \textbf{0.52} \\
& Syntactic analogies & 18.88 & 10.77 & 24.83 & 23.80 & \textbf{26.69}
& 11.84 & 24.38\\
& Semantic analogies &  8.39 & 2.84 & \textbf{25.29} & 8.45 & 19.49
& 4.50& 9.86\\
& Sentence completion &  27.69 & 21.31 & \textbf{30.18} & 25.60 & 26.89
& 25.10 & 28.88 \\
& Sentiment analysis &  74.46 & 64.85 & 70.84 & 68.48 & 71.99
& 75.51 & \textbf{75.83} \\
\hline
\hline
\multirow{5}{*}{520} & Word similarity & 0.50 & 0.31 & 0.59  & 0.53 & 0.58
& 0.58  & \textbf{0.66} \\
& Syntactic analogies & 40.67 & 22.39 & 33.49 & 52.20 & \textbf{54.64}
& 22.02 & 48.00 \\
& Semantic analogies & 28.82 & 5.37 & \textbf{62.76} &  12.58 & 39.15
& 15.46 & 41.33 \\
& Sentence completion & 30.58 & 23.11 & 33.07 & 26.69 & 26.00
&28.59& \textbf{35.86} \\
& Sentiment analysis & 81.70 & 72.97 & 78.60 & 77.38 & 79.46
&78.20& \textbf{81.90} \\
\hline
\end{tabular}
\vspace{-0.4cm}
\end{table*}

\begin{table}[t]
\caption{
Results on the syntactic and semantic analogies tasks with a bigger corpus ($M=260$).
\label{tbl:analogies}
}
\centering
\begin{tabular}{|l||r|r|r|r|}
\hline
{Task} & {CBOW}& {SG}
& \thisapproach{} \\
\hline
Syntactic & 61.37& 63.61
& \textbf{65.11}\\
Semantic & 23.13 & \textbf{54.41}
& 52.07\\
\hline
\end{tabular}
\vspace{-0.3cm}
\end{table}

\begin{table*}[t]
\caption{Comparison to previously published word representations.
The five right-most columns correspond to the tasks described above;
parenthesized values are the number of in-vocabulary items that could
be evaluated.
\label{tbl:compareothers}
}
\centering 
\begin{tabular}{|l|r|r||r|r|r|r|r|}
\hline
{Models} & $M$ & $V$ & {W.~Sim.} & {Syntactic} &{Semantic} & {Sentence} & {Sentiment} \\
\hline
\hline
{CW} & \multirow{4}{*}{50} &130,000 & \littleparen{6,225} 0.51 & \littleparen{10,427} 12.34 & \littleparen{8,656} 9.33 & \littleparen{976} 24.59 & 69.36\\
{RNN-DC} & &100,232 &   \littleparen{6,137} 0.32 & \littleparen{10,349} 10.94& \littleparen{7,853} 2.60& \littleparen{964} 19.81& 67.76\\
{HLBL} & &246,122 &  \littleparen{6,178} 0.11 & \littleparen{10,477} 8.98 & \littleparen{8,446} 1.74 & \littleparen{990} 19.90 & 62.33 \\
{NNSE} & &34,107 &  \littleparen{3,878} 0.23 & \littleparen{5,114} 1.47 & \littleparen{1,461} 2.46& \littleparen{833} 0.04 &64.80\\
{HPCA} & &178,080 & \littleparen{6,405} 0.29 & \littleparen{10,553} 10.42 & \littleparen{8,869} 3.36 & \littleparen{993} 20.14 & 67.49 \\
\hline
\hline
{\thisapproach{}} & 52 & 180,834 & \littleparen{6,525} \textbf{0.52} & \littleparen{10,675} \textbf{24.38} & \littleparen{8,733} \textbf{9.86} & \littleparen{1,004} \textbf{28.88} & \textbf{75.83}\\
\hline
\end{tabular}
\end{table*}

\paragraph{Analogies}
Unlike others tasks, our results on the syntactic and semantic analogies tasks
are below state-of-the-art performance from previous work (for all models).
We hypothesize that this is because performing well on these tasks requires training on a bigger corpus.
We combine our WMT-2011 corpus with other news corpora and Wikipedia to obtain a corpus of 6.8 billion words.
The size of the vocabulary of this corpus is 401,150.
We retrain three models that are scalable to a corpus of this size:
CBOW, SG, and \thisapproach{};\footnote{Our NCE implementation is not
  optimized and therefore not scalable.}
with $M=260$ to balance the trade-off between training time and performance ($M=52$ does not perform as well,
and $M=520$ is computationally expensive).
For \thisapproach{}, we use the fast learning algorithm in \S{\ref{sec:learning}},
since the online learning algorithm of \citet{mairal} does not scale to a problem of this size.
We report accuracies on the syntactic and semantic analogies tasks in Table~\ref{tbl:analogies}.
All models benefit significantly from a bigger corpus, and the 
performance levels are now
comparable with previous work.
On the syntactic analogies task, \thisapproach{} is the best model.
On the semantic analogies task, SG outperformed \thisapproach{}, and they both are better than CBOW.

\subsection{Other Comparisons}\label{se:found}

In Table~\ref{tbl:compareothers}, we compare with five other baseline
methods for which we do not train on our training data but pre-trained 50-dimensional word representations are available:
\begin{itemize}
\item \citet{collobert}: a neural network language model trained on Wikipedia data for 2 months (CW).\footnote{\url{http://ronan.collobert.com/senna/}}
\item \citet{huang}: a neural network model that uses additional
  global document context
  (RNN-DC).\footnote{\url{http://goo.gl/Wujc5G}
}
\item \citet{mnihhlbl}: a log bilinear model that predicts a
  word given its context, trained using hierarchical softmax (HLBL).\footnote{\url{http://metaoptimize.com/projects/wordreprs/} \citep{turian}}
\item \citet{nnse}: a word representation trained using non-negative sparse embedding (NNSE) on dependency relations and document cooccurrence counts.\footnote{Obtained from \url{http://www.cs.cmu.edu/~bmurphy/NNSE/}.}
These vectors were learned using sparse coding, but using different contexts
(dependency and document cooccurrences), a different training
method, and with a nonnegativity constraint.  Importantly, there
is no hierarchy in the code space, as in \thisapproach{}.\footnote{We
  found that NNSE trained using our contexts performed very poorly;
  see supplementary materials.}
\item \citet{hpca}: a word representation trained using Hellinger PCA (HPCA).\footnote{\url{http://lebret.ch/words/}}
\end{itemize}
These methods were all trained on different corpora, so they have
different vocabularies that do not always include all of the words
found in the tasks.  We estimate performance on the items for which
prediction is possible, and show the count for each method in
Table~\ref{tbl:compareothers}.  This comparison should be interpreted
cautiously since many experimental variables are conflated;
nonetheless, \thisapproach{} performs strongly.

\subsection{Discussion}
\label{sec:discussion}
Our method produces sparse word representations with exact zeros.
We observe that the sparse coding method without a structured regularizer produces sparser representations,
but it performs worse on our evaluation tasks, indicating that it zeroes out meaningful dimensions.  
For \thisapproach{} with $M=52$ and $M=520$, the average numbers of nonzero entries are 91\% and 85\% respectively. 
While our word representations are not extremely sparse, this makes intuitive sense since we try to represent
about 180,000 contexts in only 52 (520) dimensions.  We also did not tune $\lambda$. As we increase $M$,
we get sparser representations.

In terms of running time, \thisapproach{} is reasonably fast to learn.
We use the online dictionary learning method for $M=52$ and $M=520$ on a medium-sized corpus.
For $M=52$, the dictionary learning step took about 30 minutes (64 cores) 
and the overall learning procedure took approximately 2 hours (640 cores).
For $M=520$, the dictionary learning step took about 1.5 hours (64 cores) 
and the overall learning procedure took approximately 20 hours (640 cores).
For comparison, the SG model took about 1.5 hours and 5 hours for $M=52$ and $M=520$ using
a highly optimized implementation from the author's website (with no parallelization).
On a large corpus with 6.8 billion words and vocabulary size of about 400,000, 
\thisapproach{} with Algorithm~\ref{alg:proxgrad} took about 2 hours (16 cores) while SG took about 6.5 hours (16 cores) for $M=260$.

\begin{figure*}
\centering
\includegraphics[scale=0.63]{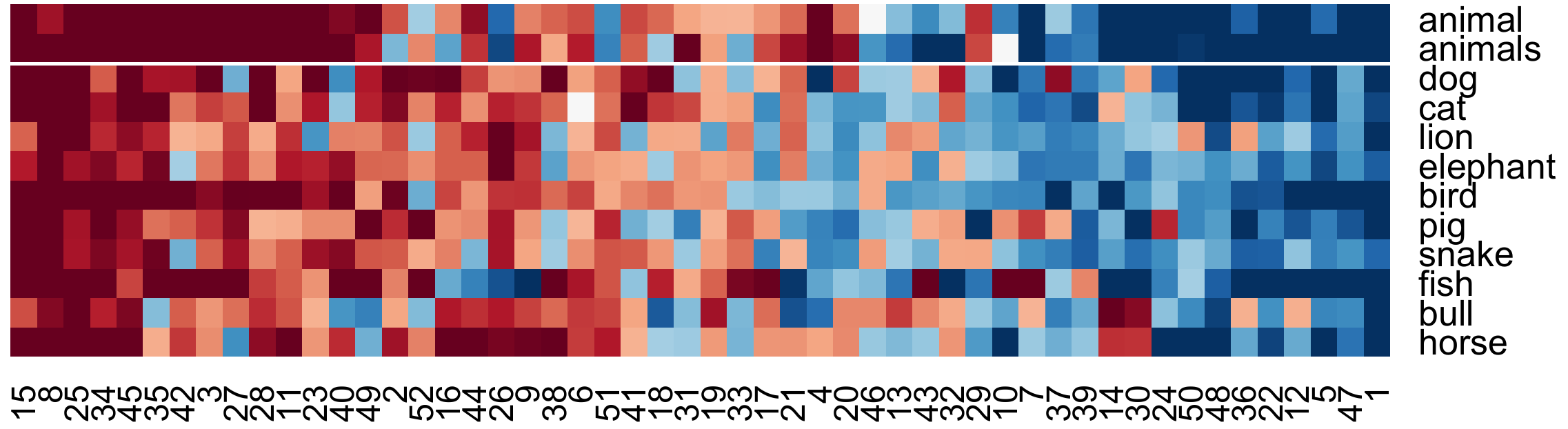}
\includegraphics[scale=0.63]{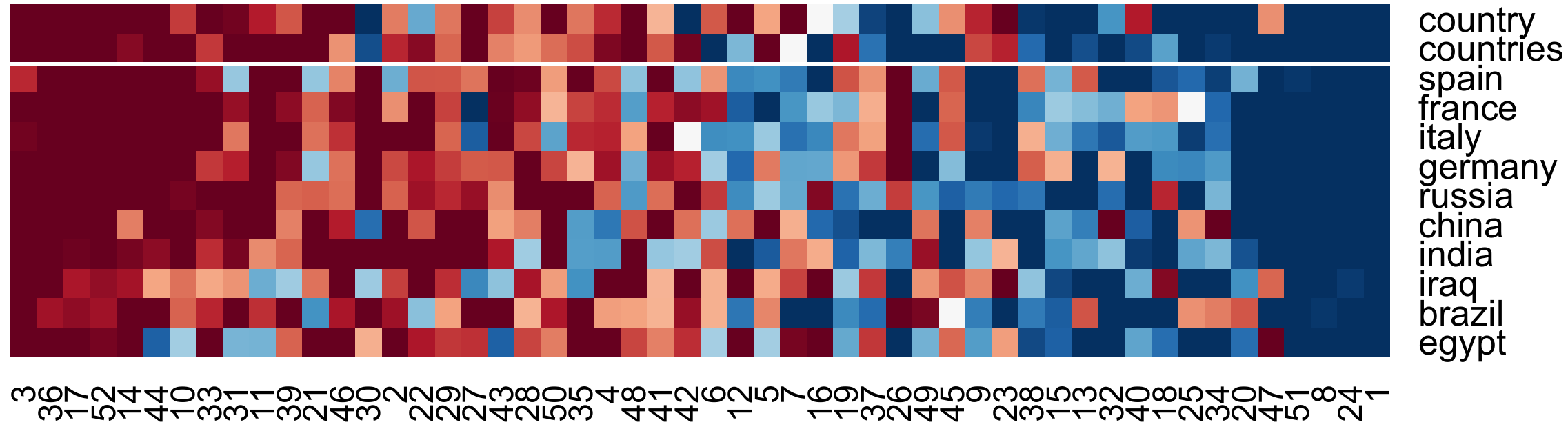}
\caption{
Heatmaps of word representations for 10 animals (top) and 10 countries (bottom) for $M=52$ from \thisapproach{}.
\textcolor{BrickRed}{Red} indicates negative values, \textcolor{RoyalBlue}{blue} indicates positive values (darker colors correspond to more extreme values);
white denotes exact zero.
The $x$-axis shows the original dimension index,
we show the dimensions from the most negative (left) to the most
positive (right), within each block, for readability.
\label{fig:patterns}
}
\end{figure*}

\begin{figure*}
\centering
\subfigure[animal]{\includegraphics[scale=0.35]{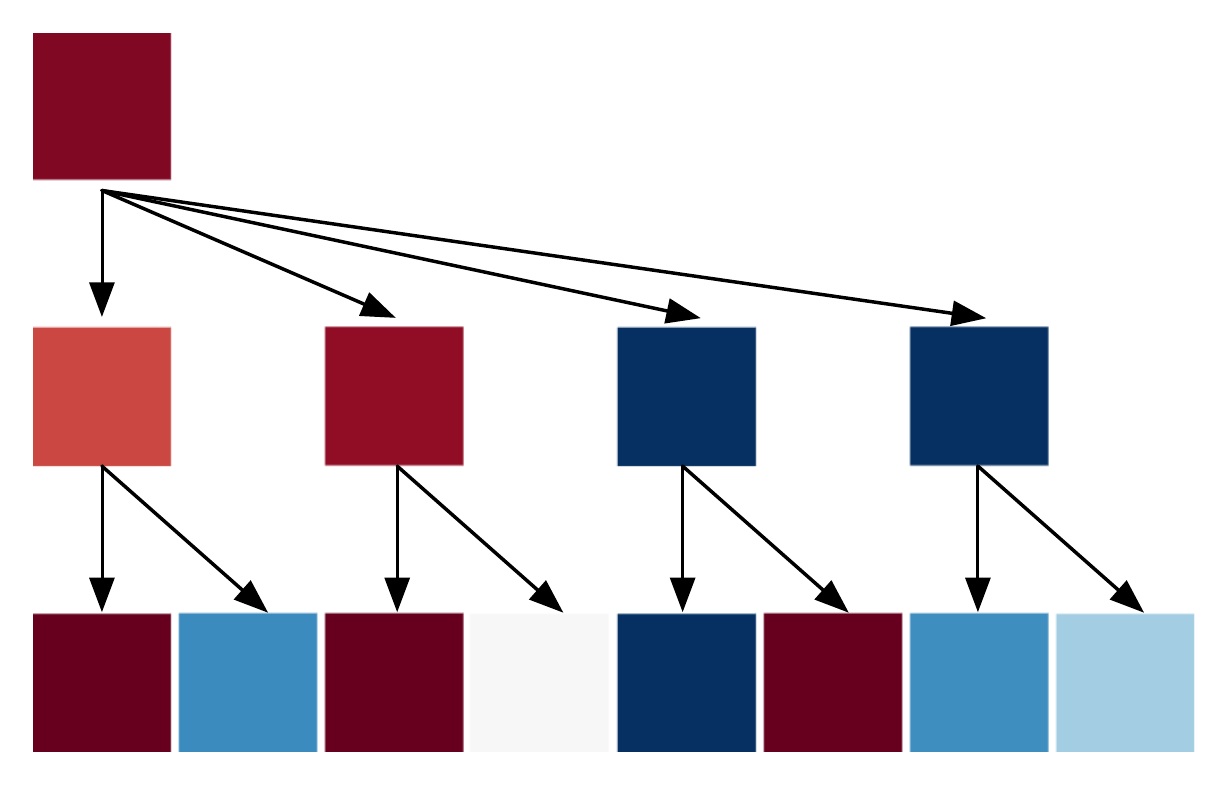}
}
\subfigure[horse]{
\includegraphics[scale=0.35]{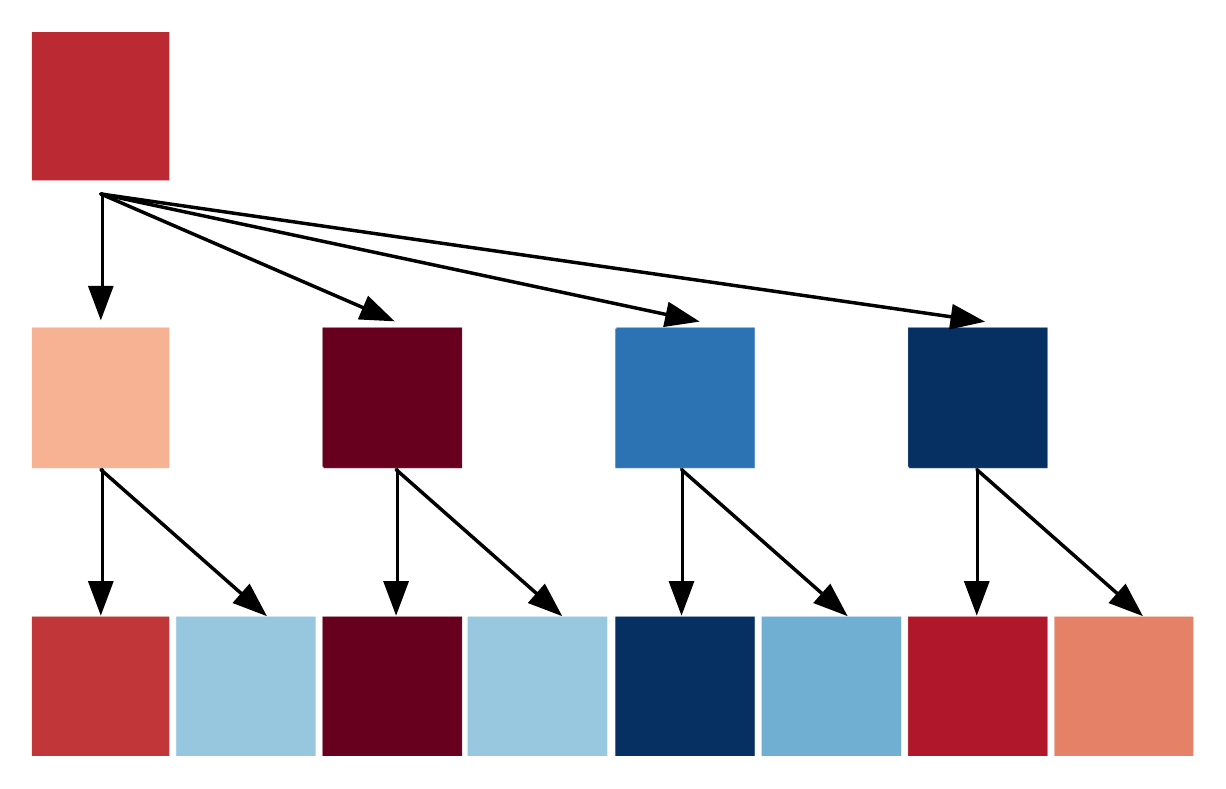}
}
\subfigure[elephant]{
\includegraphics[scale=0.35]{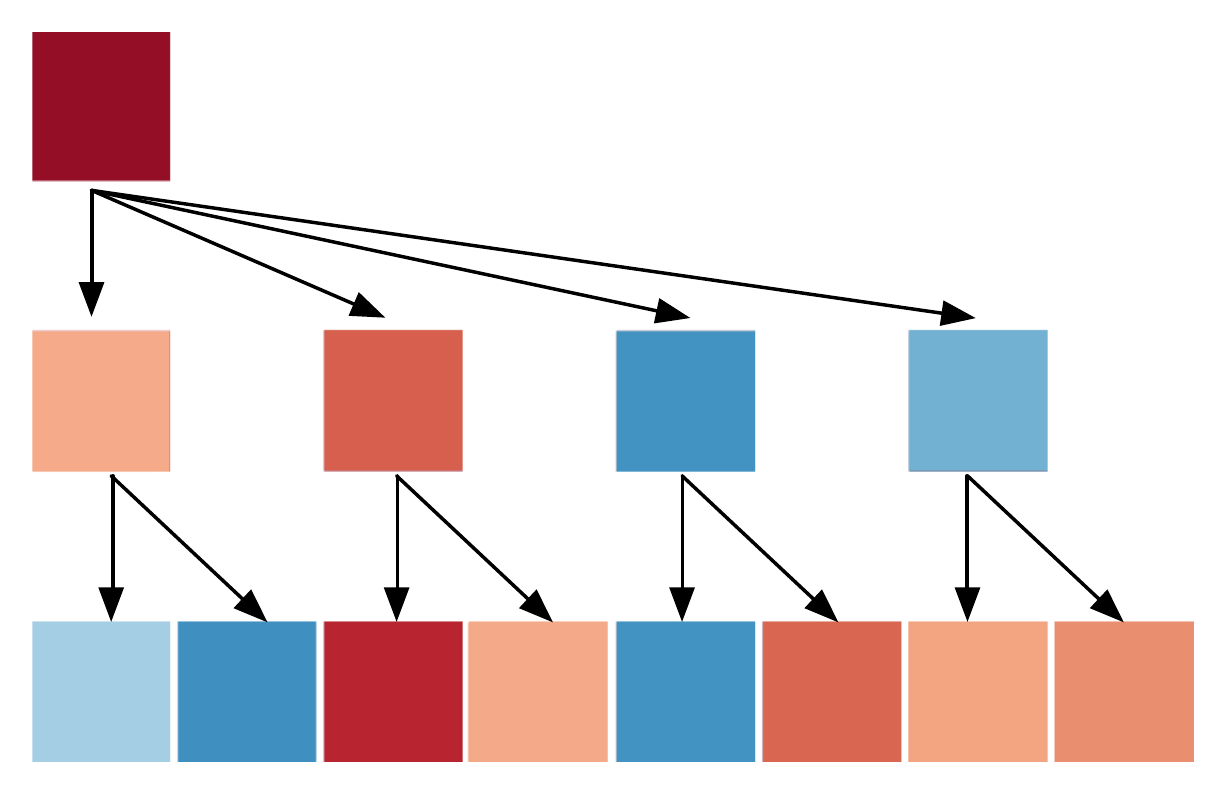}
}
 \vspace{-0.1cm}
\caption{Tree visualizations of word representations for \emph{animal} (left), \emph{horse} (center), \emph{elephant} (right)
for $M=52$.
We use the same color coding scheme as in Figure~\ref{fig:patterns}.
Here, we only show one tree (out of four), but other trees exhibit similar patterns.
\label{fig:tree_patterns}
}
 \vspace{-0.1cm}
\end{figure*}

We visualize our $M=52$ word representations (\thisapproach{}) related to animals (10 words) and countries (10 words).
We show the coefficient patterns for these words in Figure~\ref{fig:patterns}.
We can see that in both cases, there are dimensions where the coefficient signs (positive or negative) agree for all 10 words
(they are mostly on the right and left sides of the plots).
Note that the dimensions where all the coefficients agree are not the same in animals and countries.
The larger magnitude of the vectors for more abstract concepts (\emph{animal}, \emph{animals}, \emph{country}, \emph{countries}) 
is suggestive of neural imaging studies that have found evidence of more
global activation patterns for processing superordinate terms \citep{raposo}.
In Figure~\ref{fig:tree_patterns}, we show 
tree visualizations of coefficients of word representations for \emph{animal}, \emph{horse}, and \emph{elephant}.
We show one tree for $M=52$ (there are four trees in total, but other trees exhibit similar patterns).
Coefficients that differ in sign mostly correspond to leaf nodes,
validating our motivation that top level nodes should focus more on ``general''
contexts (for which they should be roughly similar for \emph{animal}, \emph{horse}, and \emph{elephant})
and leaf nodes focus on word-specific contexts. 
One of the leaf nodes for animal is driven to zero, suggesting that 
more abstract concepts require fewer dimensions to explain.  

\begin{figure*}
\centering
\includegraphics[scale=0.38]{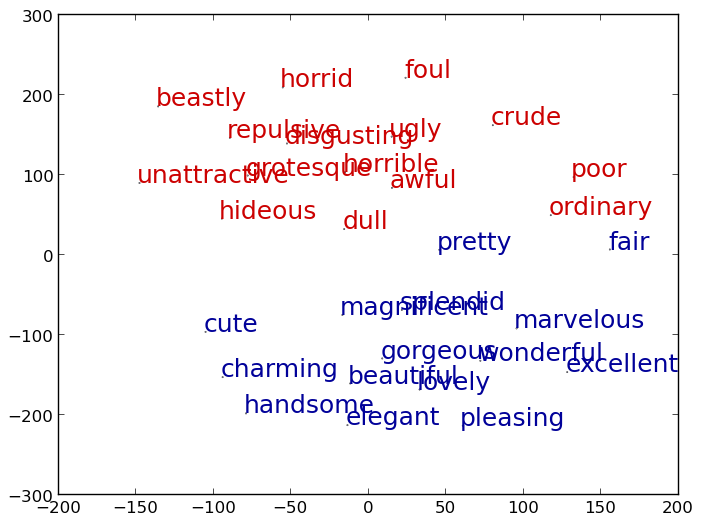}
\includegraphics[scale=0.38]{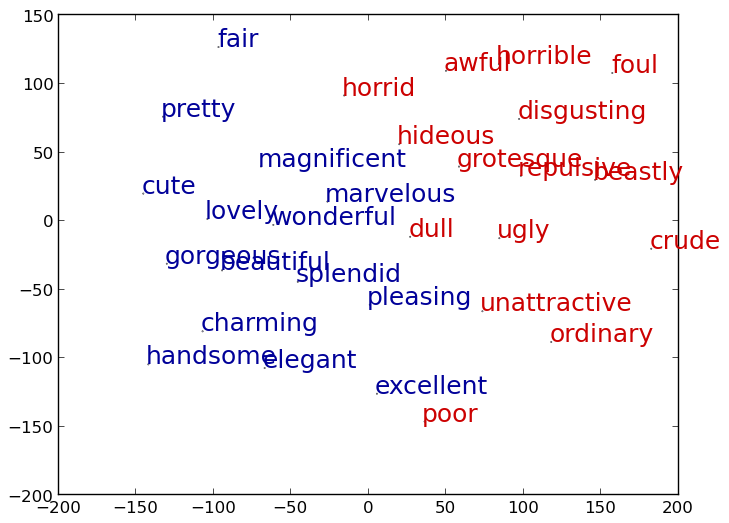}
\caption{
Two dimensional projections of the \thisapproach{} (left) and SG (right) word representations using
the t-SNE tool \citep{tsne}.
Words associated with ``good'' are colored in \textcolor{Blue}{blue},
words associated with ``bad'' are colored in \textcolor{BrickRed}{red}.
We can see that in both cases most ``good'' and ``bad'' words are clustered together (in fact, they are linearly separated in the 2D space),
except for \emph{poor} in the SG case.
See supplementary materials for more examples.
\label{fig:plot_as}
}
\end{figure*}

For \thisapproach{} and SG with $M=520$,
we project the learned word representations into two dimensions
using the t-SNE tool \citep{tsne} from \url{http://homepage.tudelft.nl/19j49/t-SNE.html}.
We show projections of words related to the concept ``good'' vs. ``bad'' in Figure~\ref{fig:plot_as}.\footnote{Since t-SNE is a non-convex method, we run it 10 times and choose the plots
with the lowest t-SNE error.} 
See supplementary materials for ``man'' vs.~``woman,'' as well as 2-dimensional projections of NCE.

\section{Conclusion}
We introduced a new method for learning word representations based on
hierarchical sparse coding.  The regularizer encourages hierarchical
organization of the latent dimensions of vector-space word embeddings.
We showed that our method outperforms state-of-the-art methods on 
word similarity ranking, syntactic analogy, sentence completion, and sentiment
analysis tasks.

\subsection*{Acknowledgements}
The authors thank anonymous reviewers, Sam Thomson, Bryan R. Routledge, Jesse Dodge, and
Fei Liu for helpful feedback on an earlier draft of this paper.
This work was supported by the National Science Foundation
through grant IIS-1352440, the Defense Advanced Research Projects Agency through grant FA87501420244, and 
computing resources provided by Google and the Pittsburgh
Supercomputing Center.

\bibliographystyle{natbib}
\bibliography{aistats2015}

\begin{thebibliography}{}

\bibitem[Bamman {\em et~al.}(2014)Bamman, Dyer, and Smith]{bamman-14}
Bamman, D., Dyer, C., and Smith, N.~A. (2014).
\newblock Distributed representations of situated language.
\newblock In {\em Proc. of ACL\/}.

\bibitem[Bengio {\em et~al.}(2003)Bengio, Ducharme, Vincent, and Jauvin]{nlm}
Bengio, Y., Ducharme, R., Vincent, P., and Jauvin, C. (2003).
\newblock A neural probabilistic language model.
\newblock {\em Journal of Machine Learning Research\/}, {\bf 3}, 1137--1155.

\bibitem[Collins and Quillian(1969)Collins and Quillian]{cogsci}
Collins, A.~M. and Quillian, M.~R. (1969).
\newblock Retrieval time from semantic memory.
\newblock {\em Journal of Verbal Learning and Verbal Behaviour\/}, {\bf 8},
  240--247.

\bibitem[Collobert {\em et~al.}(2011)Collobert, Weston, Bottou, Karlen,
  Kavukcuoglu, and Kuska]{collobert}
Collobert, R., Weston, J., Bottou, L., Karlen, M., Kavukcuoglu, K., and Kuska,
  P. (2011).
\newblock Natural language processing (almost) from scratch.
\newblock {\em Journal of Machine Learning Research\/}, {\bf 12}, 2461--2505.

\bibitem[Faruqui and Dyer(2014)Faruqui and Dyer]{manaal}
Faruqui, M. and Dyer, C. (2014).
\newblock Improving vector space word representations using multilingual
  correlation.
\newblock In {\em Proc. of EACL\/}.

\bibitem[Finkelstein {\em et~al.}(2002)Finkelstein, Gabrilovich, Matias,
  Rivlin, Solan, Wolfman, and Ruppin]{finkelstein}
Finkelstein, L., Gabrilovich, E., Matias, Y., Rivlin, E., Solan, Z., Wolfman,
  G., and Ruppin, E. (2002).
\newblock Placing search in context: The concept revisited.
\newblock {\em ACM Transactions on Information Systems\/}, {\bf 20}(1),
  116--131.

\bibitem[Fyshe {\em et~al.}(2014)Fyshe, Talukdar, Murphy, and Mitchell]{alona}
Fyshe, A., Talukdar, P.~P., Murphy, B., and Mitchell, T.~M. (2014).
\newblock Interpretable semantic vectors from a joint model of brain- and text-
  based meaning.
\newblock In {\em Proc. of ACL\/}.

\bibitem[Gutmann and Hyvarinen(2010)Gutmann and Hyvarinen]{nce}
Gutmann, M. and Hyvarinen, A. (2010).
\newblock Noise-contrastive estimation: A new estimation principle for
  unnormalized statistical models.
\newblock In {\em Proc. of AISTATS\/}.

\bibitem[Huang {\em et~al.}(2012)Huang, Socher, Manning, and Ng]{huang}
Huang, E.~H., Socher, R., Manning, C.~D., and Ng, A.~Y. (2012).
\newblock Improving word representations via global context and multiple word
  prototypes.
\newblock In {\em Proc. of ACL\/}.

\bibitem[Jenatton {\em et~al.}(2011)Jenatton, Mairal, Obozinski, and
  Bach]{hierarchicalsparsecoding}
Jenatton, R., Mairal, J., Obozinski, G., and Bach, F. (2011).
\newblock Proximal methods for hierarchical sparse coding.
\newblock {\em Journal of Machine Learning Research\/}, {\bf 12}, 2297--2334.

\bibitem[Lebret and Collobert(2014)Lebret and Collobert]{hpca}
Lebret, R. and Collobert, R. (2014).
\newblock Word embeddings through hellinger {PCA}.
\newblock In {\em Proc. of EACL\/}.

\bibitem[Lee {\em et~al.}(2007)Lee, Battle, Raina, and Ng]{honglaktrain}
Lee, H., Battle, A., Raina, R., and Ng, A.~Y. (2007).
\newblock Efficient sparse coding algorithms.
\newblock In {\em Proc. of NIPS\/}.

\bibitem[Lee {\em et~al.}(2009)Lee, Raina, Teichman, and Ng]{honglak}
Lee, H., Raina, R., Teichman, A., and Ng, A.~Y. (2009).
\newblock Exponential family sparse coding with application to self-taught
  learning.
\newblock In {\em Proc. of IJCAI\/}.

\bibitem[Luong {\em et~al.}(2013)Luong, Socher, and Manning]{luong}
Luong, M.-T., Socher, R., and Manning, C.~D. (2013).
\newblock Better word representations with recursive neural networks for
  morphology.
\newblock In {\em Proc. of CONLL\/}.

\bibitem[Mairal {\em et~al.}(2010)Mairal, Bach, Ponce, and Sapiro]{mairal}
Mairal, J., Bach, F., Ponce, J., and Sapiro, G. (2010).
\newblock Online learning for matrix factorization and sparse coding.
\newblock {\em Journal of Machine Learning Research\/}, {\bf 11}, 19--60.

\bibitem[Mikolov {\em et~al.}(2010)Mikolov, Martin, Burget, Cernocky, and
  Khudanpur]{mikolovrnn}
Mikolov, T., Martin, K., Burget, L., Cernocky, J., and Khudanpur, S. (2010).
\newblock Recurrent neural network based language model.
\newblock In {\em Proc. of Interspeech\/}.

\bibitem[Mikolov {\em et~al.}(2013a)Mikolov, Sutskever, Chen, Corrado, and
  Dean]{mikolovneg}
Mikolov, T., Sutskever, I., Chen, K., Corrado, G., and Dean, J. (2013a).
\newblock Distributed representations of words and phrases and their
  compositionality.
\newblock In {\em Proc. of NIPS\/}.

\bibitem[Mikolov {\em et~al.}(2013b)Mikolov, Chen, Corrado, and
  Dean]{mikolovsg}
Mikolov, T., Chen, K., Corrado, G., and Dean, J. (2013b).
\newblock Efficient estimation of word representations in vector space.
\newblock In {\em Proc. of ICLR Workshop\/}.

\bibitem[Miller(1995)Miller]{wordnet}
Miller, G.~A. (1995).
\newblock Wordnet: A lexical database for english.
\newblock {\em Communications of the ACM\/}, {\bf 38}(11), 39--41.

\bibitem[Mnih and Hinton(2008)Mnih and Hinton]{mnihhlbl}
Mnih, A. and Hinton, G. (2008).
\newblock A scalable hierarchical distributed language model.
\newblock In {\em Proc. of NIPS\/}.

\bibitem[Mnih and Teh(2012)Mnih and Teh]{mnihicml}
Mnih, A. and Teh, Y.~W. (2012).
\newblock A fast and simple algorithm for training neural probabilistic
  language models.
\newblock In {\em Proc. of ICML\/}.

\bibitem[Murphy {\em et~al.}(2012)Murphy, Talukdar, and Mitchell]{nnse}
Murphy, B., Talukdar, P., and Mitchell, T. (2012).
\newblock Learning effective and interpretable semantic models using
  non-negative sparse embedding.
\newblock In {\em Proc. of COLING\/}.

\bibitem[Petrov and Klein(2008)Petrov and Klein]{petrov-klein:2008:EMNLP}
Petrov, S. and Klein, D. (2008).
\newblock Sparse multi-scale grammars for discriminative latent variable
  parsing.
\newblock In {\em Proc. of EMNLP\/}.

\bibitem[Qin and Goldfarb(2012)Qin and Goldfarb]{qin}
Qin, Z.~T. and Goldfarb, D. (2012).
\newblock Structured sparsity via alternating direction methods.
\newblock {\em Journal of Machine Learning Research\/}, {\bf 13}, 1435--1468.

\bibitem[Raposo {\em et~al.}(2012)Raposo, Mendes, and Marques]{raposo}
Raposo, A., Mendes, M., and Marques, J.~F. (2012).
\newblock The hierarchical organization of semantic memory: Executive function
  in the processing of superordinate concepts.
\newblock {\em NeuroImage\/}, {\bf 59}, 1870--1878.

\bibitem[Schutze(1998)Schutze]{schutze}
Schutze, H. (1998).
\newblock Automatic word sense discrimination.
\newblock {\em Computational Linguistics - Special issue on word sense
  disambiguation\/}, {\bf 24}(1), 97--123.

\bibitem[Socher {\em et~al.}(2013)Socher, Perelygin, Wu, Chuang, Manning, Ng,
  and Potts]{socher}
Socher, R., Perelygin, A., Wu, J., Chuang, J., Manning, C., Ng, A., and Potts,
  C. (2013).
\newblock Recursive deep models for semantic compositionality over a sentiment
  treebank.
\newblock In {\em Proc. of EMNLP\/}.

\bibitem[Spearman(1904)Spearman]{spearman}
Spearman, C. (1904).
\newblock The proof and measurement of association between two things.
\newblock {\em The American Journal of Psychology\/}, {\bf 15}, 72--101.

\bibitem[Turian {\em et~al.}(2010)Turian, Ratinov, and Bengio]{turian}
Turian, J., Ratinov, L., and Bengio, Y. (2010).
\newblock Word representations: A simple and general method for semi-supervised
  learning.
\newblock In {\em Proc. of ACL\/}.

\bibitem[Turney and Pantel(2010)Turney and Pantel]{turney}
Turney, P.~D. and Pantel, P. (2010).
\newblock From frequency to meaning: Vector space models of semantics.
\newblock {\em Journal of Artificial Intelligence Research\/}, {\bf 37},
  141--188.

\bibitem[van~der Maaten and Hinton(2008)van~der Maaten and Hinton]{tsne}
van~der Maaten, L. and Hinton, G. (2008).
\newblock Visualizing data using t-sne.
\newblock {\em Journal of Machine Learning Research\/}, {\bf 9}, 2579--2605.

\bibitem[Yogatama and Smith(2014)Yogatama and Smith]{yogatama2014a}
Yogatama, D. and Smith, N.~A. (2014).
\newblock Making the most of bag of words: Sentence regularization with
  alternating direction method of multipliers.
\newblock In {\em Proc. of ICML\/}.

\bibitem[Yuan and Lin(2006)Yuan and Lin]{yuan}
Yuan, M. and Lin, Y. (2006).
\newblock Model selection and estimation in regression with grouped variables.
\newblock {\em Journal of the Royal Statistical Society, Series B\/}, {\bf
  68}(1), 49--67.

\bibitem[Zhao {\em et~al.}(2009)Zhao, Rocha, and Yu]{binyu}
Zhao, P., Rocha, G., and Yu, B. (2009).
\newblock The composite and absolute penalties for grouped and hierarchical
  variable selection.
\newblock {\em The Annals of Statistics\/}, {\bf 37}(6A), 3468--3497.

\bibitem[Zweig and Burges(2011)Zweig and Burges]{msrsentence}
Zweig, G. and Burges, C. J.~C. (2011).
\newblock The microsoft research sentence completion challenge.
\newblock Technical report, Microsoft Research Technical Report
  MSR-TR-2011-129.

\end{thebibliography}


\begin{thebibliography}{}

\bibitem[Agirre {\em et~al.}(2009)Agirre, Alfonseca, Hall, Kravalova, Pasca,
  and Soroa]{agirre}
Agirre, E., Alfonseca, E., Hall, K., Kravalova, J., Pasca, M., and Soroa, A.
  (2009).
\newblock A study on similarity and relatedness using distributional and
  wordnet-based approaches.
\newblock In {\em Proc. of NAACL-HLT\/}.

\bibitem[Bruni {\em et~al.}(2012)Bruni, Boleda, Baroni, and Tran]{bruni}
Bruni, E., Boleda, G., Baroni, M., and Tran, N.-K. (2012).
\newblock Distributional semantics in technicolor.
\newblock In {\em Proc. of ACL\/}.

\bibitem[Finkelstein {\em et~al.}(2002)Finkelstein, Gabrilovich, Matias,
  Rivlin, Solan, Wolfman, and Ruppin]{finkelstein}
Finkelstein, L., Gabrilovich, E., Matias, Y., Rivlin, E., Solan, Z., Wolfman,
  G., and Ruppin, E. (2002).
\newblock Placing search in context: The concept revisited.
\newblock {\em ACM Transactions on Information Systems\/}, {\bf 20}(1),
  116--131.

\bibitem[Halawi and Dror(2014)Halawi and Dror]{halawidror}
Halawi, G. and Dror, G. (2014).
\newblock The word relatedness mturk-771 test collection.

\bibitem[Luong {\em et~al.}(2013)Luong, Socher, and Manning]{luong}
Luong, M.-T., Socher, R., and Manning, C.~D. (2013).
\newblock Better word representations with recursive neural networks for
  morphology.
\newblock In {\em Proc. of CONLL\/}.

\bibitem[Mairal {\em et~al.}(2010)Mairal, Bach, Ponce, and Sapiro]{mairal}
Mairal, J., Bach, F., Ponce, J., and Sapiro, G. (2010).
\newblock Online learning for matrix factorization and sparse coding.
\newblock {\em Journal of Machine Learning Research\/}, {\bf 11}, 19--60.

\bibitem[Mikolov {\em et~al.}(2013)Mikolov, Chen, Corrado, and Dean]{mikolovsg}
Mikolov, T., Chen, K., Corrado, G., and Dean, J. (2013).
\newblock Efficient estimation of word representations in vector space.
\newblock In {\em Proc. of Workshop at ICLR\/}.

\bibitem[Miller and Charles(1991)Miller and Charles]{millercharles}
Miller, G.~A. and Charles, W.~G. (1991).
\newblock Contextual correlates of semantic similarity.
\newblock {\em Language and Cognitive Processes\/}, {\bf 6}(1), 1--28.

\bibitem[Murphy {\em et~al.}(2012)Murphy, Talukdar, and Mitchell]{nnse}
Murphy, B., Talukdar, P., and Mitchell, T. (2012).
\newblock Learning effective and interpretable semantic models using
  non-negative sparse embedding.
\newblock In {\em Proc. of COLING\/}.

\bibitem[Radinsky {\em et~al.}(2011)Radinsky, Agichtein, Gabrilovich, and
  Markovitch]{radinsky}
Radinsky, K., Agichtein, E., Gabrilovich, E., and Markovitch, S. (2011).
\newblock A word at a time: Computing word relatedness using temporal semantic
  analysis.
\newblock In {\em Proc. of WWW\/}.

\bibitem[Rubenstein and Goodenough(1965)Rubenstein and Goodenough]{rg}
Rubenstein, H. and Goodenough, J.~B. (1965).
\newblock Contextual correlates of synonymy.
\newblock {\em Communications of the ACM\/}, {\bf 8}(10), 627--633.

\bibitem[van~der Maaten and Hinton(2008)van~der Maaten and Hinton]{tsne}
van~der Maaten, L. and Hinton, G. (2008).
\newblock Visualizing data using t-sne.
\newblock {\em Journal of Machine Learning Research\/}, {\bf 9}, 2579--2605.

\bibitem[Yang and Powers(2006)Yang and Powers]{yangpowers}
Yang, D. and Powers, D. M.~W. (2006).
\newblock Verb similarity on the taxonomy of wordnet.
\newblock In {\em Proc. of GWC\/}.

\end{thebibliography}
\end{document}


\maketitle

\section{Additional Results}
In Table~\ref{tbl:supresults}, we compare \thisapproach{} with three additional baselines:
\begin{itemize}
\item \citet{nnse}: a word representation trained using non-negative sparse embedding (NNSE) on our corpus. Similar to the authors, we use an NNSE implementation from \url{http://spams-devel.gforge.inria.fr/} \citep{mairal}.
\item \citet{mikolovsg}: a log bilinear model that predicts a word given its context, trained using hierarchical softmax with a binary Huffman tree (continuous bag of words, CBOW-HS). We use an implementation from \url{https://code.google.com/p/word2vec/}.
\item \citet{mikolovsg}: a log bilinear model that predicts context words given a target word, trained using hierarchical softmax with a binary Huffman tree (skip gram, SG-HS). We use an implementation from \url{https://code.google.com/p/word2vec/}.
\end{itemize}
We train these models on our corpus using the same setup as experiments in our paper.

\begin{table*}[h]
\caption{
Summary of results for non-negative sparse embedding (NNSE),
continuous bag-of-words and skip gram models trained with hierarchical softmax (CBOW-HS and SG-HS).
Higher number is better (higher correlation coefficient or higher accuracy).
\label{tbl:supresults}
}
\centering
\begin{tabular}{|c|l||r|r|r||r|}
\hline
{$M$} & {Task} & {NNSE} & {CBOW-HS} & {SG-HS} & \thisapproach{}\\
\hline
\hline
\multirow{5}{*}{52} & Word similarity & 0.04 & 0.38 & 0.47 & \textbf{0.52} \\
& Syntactic analogies & 0.10 & 19.50 & \textbf{24.87} & 24.38 \\ 
& Semantic analogies & 0.01 & 5.31 & \textbf{14.77} & 9.86 \\
& Sentence completion & 0.01 & 22.51 & 28.78 & \textbf{28.88} \\ 
& Sentiment analysis & 61.12 & 68.92 & 71.72 & \textbf{75.83} \\
\hline
\hline
\multirow{5}{*}{520} & Word similarity & 0.05 & 0.50 & 0.57 & \textbf{0.66} \\ 
& Syntactic analogies & 0.81 & 46.00 & \textbf{50.40} & 48.00 \\
& Semantic analogies & 0.57 & 8.00 & 31.05 & \textbf{41.33} \\
& Sentence completion &  22.81 & 25.80 & 27.79 & \textbf{35.86} \\
& Sentiment analysis & 67.05 & 78.50 & 79.57 & \textbf{81.90} \\
\hline
\end{tabular}
\vspace{0.3cm}
\end{table*}

\section{Additional Two Dimensional Projections}
\begin{figure*}[t]
\includegraphics[scale=0.4]{mine_as.png}
s\includegraphics[scale=0.4]{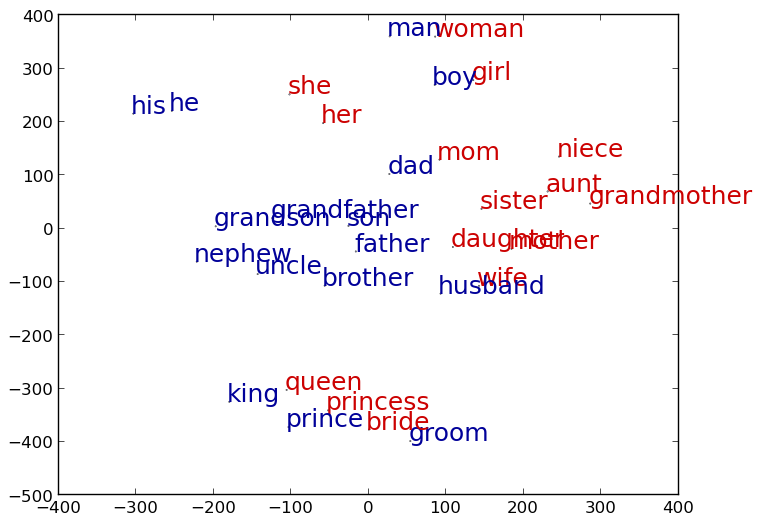} \\
\includegraphics[scale=0.4]{sg_neg_as.png} 
\includegraphics[scale=0.4]{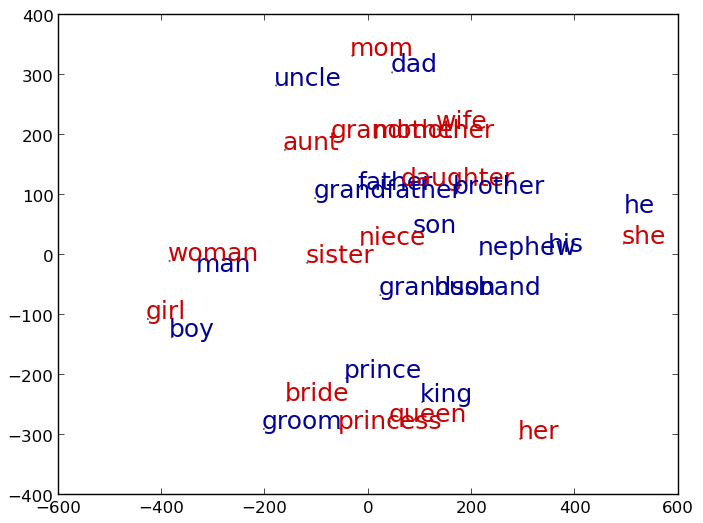} \\
\includegraphics[scale=0.4]{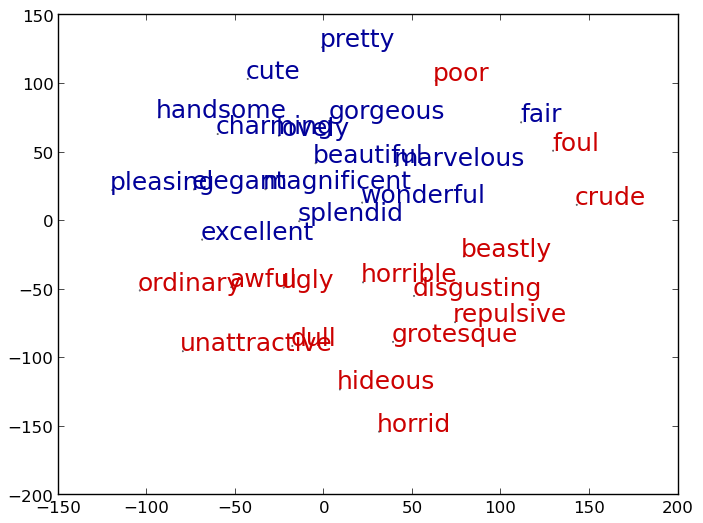} 
\includegraphics[scale=0.4]{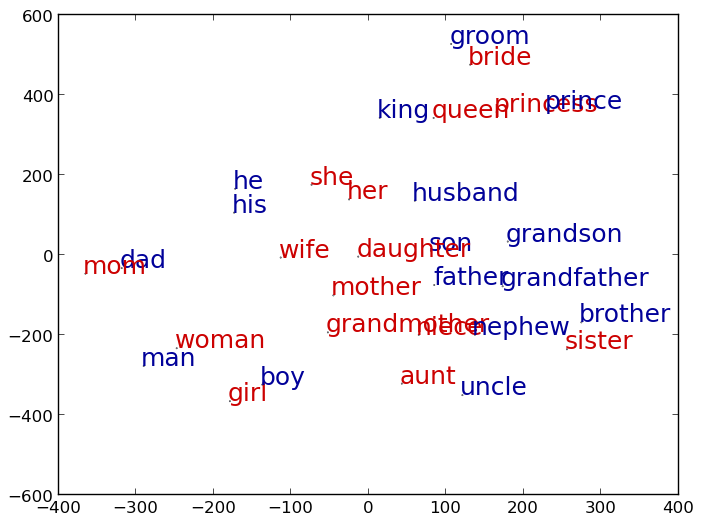}
\caption{
Two dimensional projections of the \thisapproach{} (top), SG (middle), and NCE (bottom) word representations using
the t-SNE tool \citep{tsne}.
Words associated with ``good'' (left) and ``man'' (right) are colored in \textcolor{Blue}{blue},
words associated with ``bad'' (left) and ``woman'' (right) are colored in \textcolor{BrickRed}{red}.
The two plots on the top left are the same plots shown in the paper.
\label{fig:plot_as_mf}
}
\end{figure*}

For \thisapproach{}, SG, and NCE with $M=520$,
we project the learned word representations into two dimensions 
using the t-SNE tool \citep{tsne} from \url{http://homepage.tudelft.nl/19j49/t-SNE.html}.
We show projections of words related to the concept ``good'' vs. ``bad'' and ``man'' vs. ``woman'' in Figure~\ref{fig:plot_as_mf}.

\section{List of Word Similarity Datasets}
We use the following word similarity datasets in our experiments:
\begin{itemize}
\item \citet{finkelstein}: WordSimilarity dataset (353 pairs).
\item \citet{agirre}: a subset of WordSimilarity dataset for evaluating similarity (203 pairs).
\item \citet{agirre}: a subset of WordSimilarity dataset for evaluating relatedness (252 pairs).
\item \citet{millercharles}: semantic similarity dataset (30 pairs)
\item \citet{rg}: contains only nouns (65 pairs)
\item \citet{luong}: rare words (2,034 pairs)
\item \citet{bruni}: frequent words (3,000 pairs)
\item \citet{radinsky}: MTurk-287 dataset (287 pairs)
\item \citet{halawidror}: MTurk-771 dataset (771 pairs)
\item \citet{yangpowers}: contains only verbs (130 pairs)
\end{itemize}

\bibliographystyle{natbib}
\bibliography{supplementary}